\documentclass{article}

\PassOptionsToPackage{numbers, compress}{natbib}
\usepackage[final]{nips_2017}

\usepackage[utf8]{inputenc} 
\usepackage[T1]{fontenc}    
\usepackage{hyperref}       
\usepackage{url}            
\usepackage{booktabs}       
\usepackage{amsfonts}       
\usepackage{nicefrac}       
\usepackage{microtype}      

\def\vec#1{\mathbf{#1}}
\usepackage{bm}
\usepackage{amsmath}
\usepackage{amssymb}
\usepackage{amsfonts}
\usepackage{graphicx}
\usepackage{natbib}

\title{Improving Output Uncertainty Estimation\\and Generalization in Deep Learning\\via Neural Network Gaussian Processes} 

\author{Tomoharu Iwata\\
NTT Communication Science Laboratories\\
\texttt{iwata.tomoharu@lab.ntt.co.jp}\\
\AND Zoubin Ghahramani\\
University of Cambridge, Uber AI Labs\\
\texttt{zoubin@eng.cam.ac.uk}}

\usepackage{version}	
\begin{document}

\maketitle

\begin{abstract}
We propose a simple method that combines neural networks and Gaussian processes.
The proposed method can estimate the uncertainty of outputs and flexibly adjust target functions where training data exist, which are advantages of Gaussian processes. The proposed method can also achieve high generalization performance for unseen input configurations, which is an advantage of neural networks. With the proposed method, neural networks are used for the mean functions of Gaussian processes. We present a scalable stochastic inference procedure, where sparse Gaussian processes are inferred by stochastic variational inference, and the parameters of neural networks and kernels are estimated by stochastic gradient descent methods, simultaneously. We use two real-world spatio-temporal data sets to demonstrate experimentally that the proposed method achieves better uncertainty estimation and generalization performance than neural networks and Gaussian processes.
\end{abstract}

\section{Introduction}

Neural networks (NNs) have achieved state-of-the-art results
in a wide variety of supervised learning tasks, 
such as image recognition~\citep{krizhevsky2012imagenet,Simonyan14c,russakovsky2015imagenet}, 
speech recognition~\citep{seide2011conversational,hinton2012deep,dahl2012context} 
and machine translation~\citep{bahdanau2014neural,cholearning}.
However, NNs have a major drawback in that output uncertainty is not well estimated.
NNs give point estimates of outputs at test inputs.

Estimating the uncertainty of the output is important in various situations.
First, the uncertainty can be used for rejecting the results.
In real-world applications such as medical diagnosis,
we should avoid automatic decision making with the difficult examples,
and ask human experts or conduct other examinations to achieve high reliability.
Second, the uncertainty can be used to calculate risk.
In some domains, it is important to be able to estimate the probability of 
critical issues occurring, 
for example with self-driving cars or nuclear power plant systems.
Third, the uncertainty can be used for the inputs of other machine learning tasks.
For example, uncertainty of speech recognition results helps in terms of 
improving 
machine translation performance in automatic speech translation systems~\citep{ney1999speech}.
The uncertainty would also be helpful for active learning~\citep{krause2007nonmyopic} and reinforcement learning~\citep{blundell2015weight}.

We propose a simple method that makes it possible for NNs to estimate output uncertainty.
With the proposed method, NNs are used 
for the mean functions of Gaussian processes 
(GPs)~\citep{rasmussen2006gaussian}.
GPs are used as prior distributions over smooth nonlinear functions,
and the uncertainty of the output can be estimated with Bayesian inference.
GPs perform well in various regression and classification tasks~\citep{williams1996gaussian,barber1997gaussian,naish2007generalized,nickisch2008approximations}.

Combining NNs and GPs gives us another advantage.
GPs exploit local generalization,
where generalization is achieved by 
local interpolation between neighbors~\citep{bengio2013representation}.
Therefore, GPs can adjust target functions rapidly in the presense of training data, but fail to generalize in regions where there are no training data.
On the other hand, 
NNs have good generalization capability for unseen input configurations 
by learning multiple levels of distributed representations,
but require a huge number of training data.
Since GPs and NNs achieve generalization in different ways,
the proposed method can improve generalization performance
by adopting both of their advantages.

\begin{figure}[t]
\centering
{\tabcolsep=-0.5em
\begin{tabular}{cccc}
no training & trained w/ two samples & 
no training & trained w/ two samples \\ 
\includegraphics[width=11em]{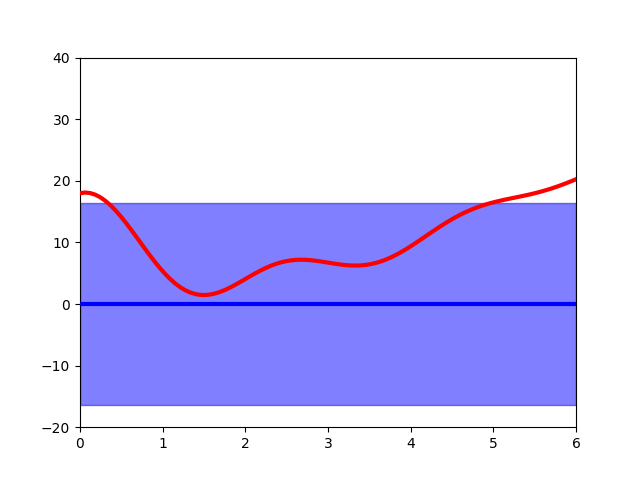}&
\includegraphics[width=11em]{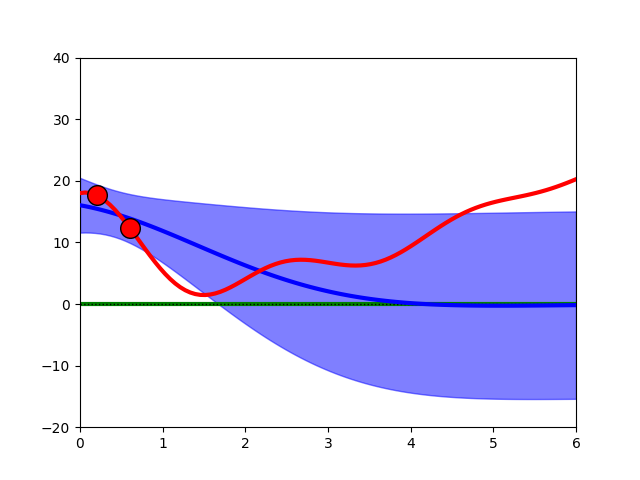}&
\includegraphics[width=11em]{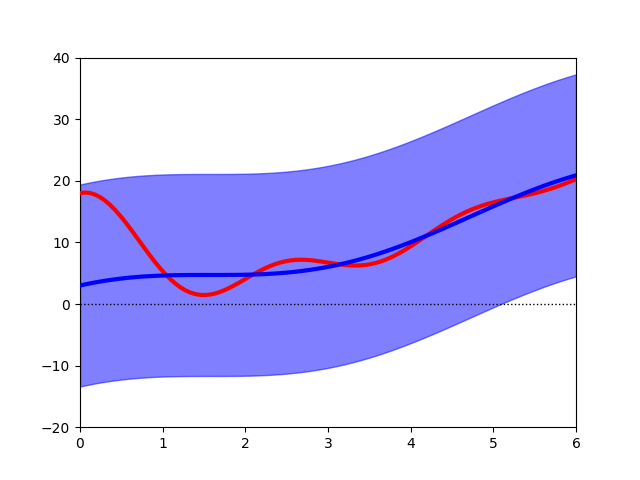}&
\includegraphics[width=11em]{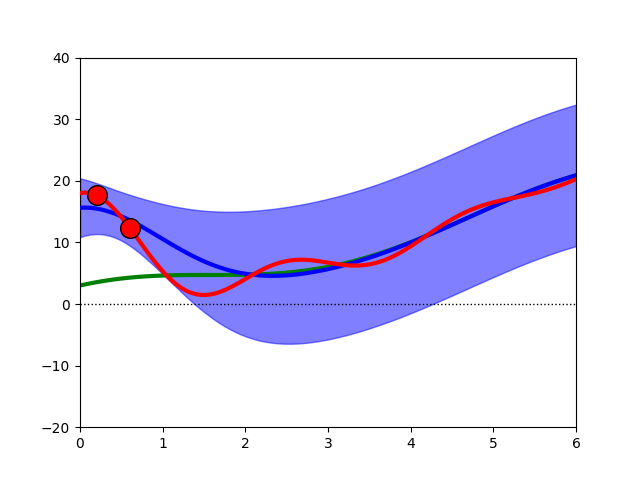}\\
\multicolumn{2}{c}{(a) GP with zero mean functions} &
\multicolumn{2}{c}{(b) GP with nonzero mean functions}\\
\end{tabular}}
\caption{True values (red), mean function values (green) and prediction values (blue) provided by GPs with zero mean functions (a) and GPs with nonzero mean functions (b). The blue area is the 95\% confidence interval of the prediction, and the red points indicate the training samples.}
\label{fig:GPillustration}
\end{figure}

Zero mean functions are usually used
since GPs with zero mean functions and some specific kernels can approximate an arbitrary continuous function 
given enough training data~\citep{micchelli2006universal}.
However, GPs with zero mean functions predict zero outputs far from training samples.
Figure~\ref{fig:GPillustration}(a) shows the predictions of GPs with zero mean functions and RBF kernels.
When trained with two samples, the prediction values 
are close to the true values if there are training samples, 
but far from the true values if there are none.
On the other hand, when GPs with appropriate nonzero mean functions are used as in Figure~\ref{fig:GPillustration}(b),
the prediction approximates the true values 
even when there are no training samples.
Figure~\ref{fig:GPillustration} shows that GPs rapidly adjust the prediction 
when there are training data
regardless of the mean function values.

The proposed method gives NNs more flexibility via GPs.
In general, the risk of overfitting increases as the model flexibility increases.
However, since the proposed method is based on Bayesian inference,
where nonlinear functions with GP priors are integrated out,
the proposed method can help alleviate overfitting.

To retain the high generalization capability of NNs with the proposed method,
large training data are required.
The computational complexity of the exact inference of GPs is
cubic in the number of training samples, which is prohibitive for large data.
We present a scalable stochastic inference procedure for the proposed method,
where sparse GPs are inferred by stochastic variational inference~\citep{hensman2013gaussian},
and NN parameters and kernel parameters are estimated by stochastic gradient descent methods, simultaneously.
By using stochastic optimization, the parameters are updated efficiently without analyzing all the data at each iteration,
where a noisy estimate of the gradient of the objective function is used.
The inference algorithm also enables us to handle massive data even when they cannot be stored in a memory.

\section{Related work}
\label{sec:related}

Bayesian NNs are the most common way of introducing uncertainty into NNs,
where distributions over the NN parameters are inferred.
A number of Bayesian NN methods have been proposed including 
Laplace approximation~\citep{mackay1992practical},
Hamiltonian Monte Carlo~\citep{neal1995bayesian},
variational inference~\citep{hinton1993keeping,graves2011practical,blundell2015weight,Louizos2016,sun2017learning},
expectation propagation~\citep{jylanki2014expectation},
stochastic backpropagation~\citep{hernandez2015probabilistic}, 
and
dropout~\citep{kingma2015variational,gal2016dropout} methods.
Our proposed method gives the output uncertainty of 
NNs with a different approach,
where we conduct point estimation for the NN parameters, 
but the NNs are combined with GPs.
Therefore, the proposed method incorporates the high generalization performance of NNs and the high flexibility of GPs,
and can handle large-scale data by using scalable NN stochastic optimization and GP stochastic variational inference.

Although zero mean functions are often used for GPs, 
nonzero mean functions, such as polynomial functions~\citep{blight1975bayesian}, have also been used.
When the mean functions are linear in the parameters, the parameters can be integrated out,
which leads to another GP~\citep{o1978curve}.
However, scalable inference algorithms for GPs with flexible nonlinear mean functions like NNs have not been proposed.

NNs and GPs are closely related. NNs with a hidden layer converge to GPs in the limit of an infinite number of hidden units~\citep{neal1995bayesian}.
A number of methods combining NNs and GPs have been proposed.
Deep GPs~\citep{damianou2013deep} use GPs for each layer in NNs, where local generalization is exploited since their inputs are kernel values.
GP regression networks~\citep{wilson2012gaussian} 
combine the structural properties of NNs 
with the nonparametric flexibility of GPs 
for accommodating input dependent signal and noise correlations.
Manifold GPs~\citep{calandra2016manifold} and deep NN based GPs~\citep{huang2015scalable} use NNs for transforming the input features of GPs.
Deep kernel learning~\citep{wilson2016deep} uses NNs to learn kernels for GPs.
The proposed method is different from these methods since it incorporates 
the outputs of NNs into GPs.

\section{Proposed method}
\label{sec:method}

Suppose that we have a set of input and output pairs,
${\cal D}=(\mathbf{x}_{n},y_{n})_{n=1}^{N}$,
where $\mathbf{x}_{n}\in\mathbb{R}^{D}$ is the $n$th input,
and $y_{n}\in\mathbb{R}$ is its output.
Output $y_{n}$ is assumed to be generated by a nonlinear function
$f(\mathbf{x}_{n})$ with Gaussian noise.
Let $\mathbf{f}=(f_{n})_{n=1}^{N}$ be the vector of function values on
the observed inputs, $f_{n}=f(\mathbf{x}_{n})$.
Then, the probability of the output 
$\mathbf{y}=(\mathbf{y}_{n})_{n=1}^{N}$ 
is given by
\begin{align}
p(\mathbf{y}|\mathbf{f})=\prod_{n=1}^{N}{\cal N}(y_{n}|f_{n},\beta^{-1}),
\end{align}
where $\beta$ is the observation precision parameter.
For the nonlinear function, we assume a GP model,
\begin{align}
f(\mathbf{x}) \sim {\cal GP}(g(\mathbf{x};\bm{\phi}),k(\mathbf{x},\mathbf{x}';\bm{\theta})),
\label{eq:nngp}
\end{align}
where $g(\mathbf{x};\bm{\phi})$ is the mean function 
with parameters $\bm{\phi}$,
and $k(\mathbf{x},\mathbf{x}';\bm{\theta})$ is the kernel function
with kernel parameters $\bm{\theta}$.
We use a NN for the mean function, and
we call (\ref{eq:nngp}) NeuGaP,
which is a simple and new model that fills 
a gap between the GP and NN literatures.
By integrating out the nonlinear function $\vec{f}$,
the likelihood is given by
\begin{align}
p(\vec{y}|\vec{X},\bm{\phi},\bm{\theta})=
{\cal N}(\vec{y}|\vec{g},\vec{K})=
(2\pi)^{-\frac{N}{2}}|\vec{K}|^{-\frac{1}{2}}
\exp\left(-\frac{1}{2}(\vec{y}-\vec{g})^{\top}\vec{K}^{-1}(\vec{y}-\vec{g})
\right),
\label{eq:gplikelihood}
\end{align}
where $\vec{X}=(\vec{x}_{n})_{n=1}^{N}$, 
$\vec{K}$ is the $N\times N$ covariance matrix defined by the kernel function
$k(\vec{x},\vec{x}';\bm{\theta})$,
and
$\vec{g}=(g(\vec{x}_{n}))_{n=1}^{N}$
is the vector of the output values of the NN on the observed inputs.
The parameters in GPs are usually estimated 
by maximizing the marginal likelihood (\ref{eq:gplikelihood}).
However, the exact inference is infeasible for large data since 
the computational complexity is $O(N^{3})$
due to the inversion of the covariance matrix.


To reduce the computational complexity
while keeping the desirable properties of GPs,
we employ sparse GPs~\citep{snelson2006sparse,quinonero2005unifying,hensman2013gaussian}.
With a sparse GP,
inducing inputs $\mathbf{Z}=(\mathbf{z}_{m})_{m=1}^{M}$,
$\vec{z}_{m}\in\mathbb{R}^{D}$,
and their outputs $\mathbf{u}=(u_{m})_{m=1}^{M}$,
$u_{m}\in\mathbb{R}$, are introduced.
The basic idea behind sparse inducing point methods is that when the number of inducing points $M \ll N$,
computation can be reduced in $O(M^2N)$.
The inducing outputs $\vec{u}$ are assumed to be
generated by the nonlinear function of NeuGaP (\ref{eq:nngp})
taking the inducing inputs $\vec{Z}$ as inputs.
By marginalizing out the nonlinear function,
the probability of the inducing outputs is given by
\begin{align}
p(\vec{u})={\cal N}(\vec{u}|\vec{g}_{M},\vec{K}_{MM}),
\end{align}
where $\vec{g}_{M}=(g(\mathbf{z}_{m}))_{m=1}^{M}$ is the vector
of the NN output values on the inducing inputs,
$\vec{K}_{MM}$ is the $M\times M$ covariance matrix evaluated 
between all the inducing inputs, 
$\vec{K}_{MM}(m,m')=k(\mathbf{z}_{m},\mathbf{z}_{m'})$.
The output values at the observed inputs $\mathbf{f}$ are assumed to be 
conditionally independent of each other given the inducing outputs $\mathbf{u}$,
then we have
\begin{align}
p(\vec{f}|\vec{u})=
\prod_{n=1}^{N}p(f_{n}|\vec{u})
=\prod_{n=1}^{N}{\cal N}(f_{n}|\mu_{n},\tilde{k}_{n}),
\label{eq:fu}
\end{align}
where 
\begin{align}
\mu_{n}=g(\vec{x}_{n})+\vec{k}_{Mn}^{\top}\vec{K}_{MM}^{-1}(\vec{u}-\vec{g}_{M}), \quad
\tilde{k}_{n} = k(\vec{x}_{n},\vec{x}_{n})-\vec{k}_{Mn}^{\top}\vec{K}_{MM}^{-1}\vec{k}_{Mn}.
\end{align}
Here, $\vec{k}_{Mn}$ is the $M$-dimensional column vector of the covariance function
evaluated between observed and inducing inputs, 
$\vec{k}_{Mn}(m)=k(\mathbf{x}_{n},\mathbf{z}_{m})$.
Equation (\ref{eq:fu}) is obtained in the same way as
the derivation of the predictive mean and variance of test data points 
in standard GPs.


The lower bound of the log marginal likelihood of 
the sparse GP to be maximized is
\begin{align}
\log p(\vec{y}) &= \log \int p(\vec{y}|\vec{u})p(\vec{u})d\vec{u}
\geq
\int q(\vec{u}) 
\log p
(\vec{y}|\vec{u})\frac{p(\vec{u})}{q(\vec{u})}
d\vec{u},
\label{eq:py}
\end{align}
where $q(\vec{u})={\cal N}(\vec{m},\vec{S})$
is the variational distribution of the inducing points,
and Jensen's inequality is applied~\citep{titsias2009variational}.
The log likelihood of the observed output $\vec{y}$
given the inducing points $\mathbf{u}$ is as follows,
\begin{align}
\log p(\mathbf{y}|\vec{u}) &=\log \int p(\mathbf{y}|\mathbf{f})p(\mathbf{f}|\mathbf{u})d\mathbf{f}
\geq \int  p(\mathbf{f}|\mathbf{u}) \log p(\mathbf{y}|\mathbf{f}) d\mathbf{f}
= \sum_{n=1}^{N} \log {\cal N}(y_{n}|\mu_{n},\beta^{-1})-\frac{1}{2}\beta \tilde{k}_{n},
\label{eq:pyu}
\end{align}
where Jensen's inequality is applied, and 
the lower bound of $\log p(\vec{y}|\vec{u})$ is decomposed into 
terms for each training sample.
By using (\ref{eq:py}) and (\ref{eq:pyu}), 
the lower bound of $\log p(\vec{y})$ is given by
\begin{align}
\log p(\vec{y}) &\geq 
\sum_{n=1}^{N} \left(\log {\cal N}(y_{n}|\tilde{\mu}_{n},\beta^{-1})-\frac{1}{2}\beta\tilde{k}_{n}
-\frac{1}{2}{\rm tr}(\vec{S}\bm{\Lambda}_{n})\right)
-{\rm KL}(q(\vec{u})||p(\vec{u}))\equiv L,
\label{eq:lowerbound}
\end{align}
where 
\begin{align}
\tilde{\mu}_{n}=g(\vec{x}_{n})+\vec{k}_{Mn}^{\top}\vec{K}_{MM}^{-1}(\vec{m}-\vec{g}_{M}), \quad
\bm{\Lambda}_{n}=\beta\vec{K}_{MM}^{-1}\vec{k}_{Mn}\vec{k}_{Mn}^{\top}\vec{K}_{MM}^{-1},
\end{align}
and ${\rm KL}(q(\vec{u})||p(\vec{u}))$ is the KL divergence
between two Gaussians, which is calculated by 
\begin{align}
{\rm KL}(q(\vec{u})||p(\vec{u}))
=\frac{1}{2}\left(
\log\frac{|\vec{K}_{MM}|}{|\vec{S}|}
-M+{\rm tr}(\vec{K}_{MM}^{-1}\vec{S})+(\vec{m}-\vec{g}_{M})^{\top}\vec{K}_{MM}^{-1}(\vec{m}-\vec{g}_{M})
\right).
\end{align}

The NN parameters $\bm{\phi}$ and kernel parameters $\bm{\theta}$
are updated efficiently by maximizing the lower bound (\ref{eq:lowerbound})
using stochastic gradient descent methods.
The parameters in the variational distribution, $\vec{m}$ and $\vec{S}$,
are updated efficiently by using stochastic variational inference~\citep{hoffman2013stochastic}.
We altenately iterate the stochastic gradient descent and stochastic variational inference for each minibatch of training data.

With stochastic variational inference,
the parameters of variational distributions are updated based on the natural gradients~\citep{amari1998natural},
which are computed by multiplying the gradients by the inverse of the Fisher information matrix.
The natural gradients provide faster convergence than standard gradients 
by taking account of the information geometry of the parameters.
In the exponential family, the natural gradients with respect to natural parameters
correspond to the gradients with respect to expectation parameters~\citep{hensman2012fast}.
The natural parameters of Gaussian ${\cal N}(\vec{m},\vec{S})$ are
$\bm{\lambda}_{1}=\vec{S}^{-1}\vec{m}$ and
$\bm{\lambda}_{2}=-\frac{1}{2}\vec{S}^{-1}$.
Its expectation parameters are
$\bm{\eta}_{1}=\vec{m}$
and
$\bm{\eta}_{2}=\vec{m}\vec{m}^{\top}+\vec{S}$.
We take a step in the natural gradient direction by employing
$\bm{\lambda}^{(t+1)}=\bm{\lambda}^{(t)}+\ell_{t} \frac{\partial L}{\partial \bm{\eta}}$,
where $\frac{\partial L}{\partial \bm{\eta}}=G(\bm{\lambda})^{-1}\frac{\partial L}{\partial \bm{\lambda}}$ is the natural gradient of the objective function with respect to the natural parameter,
$G(\bm{\lambda})$ is the Fisher information, and $\ell_{t}$ is the step length at iteration $t$.
The update rules for the proposed model are given by
\begin{align}
\bm{\lambda}_{1}^{(t+1)}
=
\ell_{t} \Bigl(\beta\vec{K}_{MM}^{-1}\vec{k}_{Mn}(y_{n}-g_{n}+\vec{k}_{Mn}^{\top}\vec{K}_{MM}^{-1}\vec{g}_{M})\Bigr)
+(1-\ell_{t})\bm{\lambda}_{1}^{(t)},
\label{eq:svi1}
\end{align}
\begin{align}
\bm{\lambda}_{2}^{(t+1)}=
\ell_{t} \left( -\frac{1}{2}(
\bm{\Lambda}_{n}
+\vec{K}_{MM}^{-1})\right)
+ (1-\ell_{t}) \bm{\lambda}_{2}^{(t)}.
\label{eq:svi2}
\end{align}
We can use minibatches instead of a single training sample
to update the natural parameters.

The output distribution 
given test input $\vec{x}^{*}$ is calculated by
\begin{align}
p(y^{*}|\vec{x}^{*},{\cal D})&\approx
\int\int p(y^{*}|f^{*})p(f^{*}|\vec{u})q(\vec{u})df^{*}d\vec{u}\nonumber\\
&={\cal N}(y^{*}|g(\vec{x}^{*})+\vec{k}_{M*}^{\top}\vec{K}_{MM}^{-1}(\vec{m}-\vec{g}_{M}),\beta^{-1}+\tilde{k}_{*}+\vec{k}_{M*}^{\top}\vec{K}_{MM}^{-1}\vec{S}\vec{K}_{MM}^{-1}\vec{k}_{M*}),
\end{align}
where $\vec{k}_{*M}$ is the covariance function column vector evaluated 
between test input $\vec{x}^{*}$ and inducing inputs,
and $\tilde{k}_{*} = k(\vec{x}_{*},\vec{x}_{*})-\vec{k}_{M*}^{\top}\vec{K}_{MM}^{-1}\vec{k}_{M*}$.

\section{Experiments}
\label{sec:experiment}

\paragraph{Data}

We evaluated our proposed method by using two real-world spatio-temporal 
data sets.
The first data set is the Comprehensive Climate (CC) data set
\footnote{\url{http://www-bcf.usc.edu/~liu32/data/NA-1990-20002-Monthly.csv}},
which consists of monthly climate reports for North America~\citep{bahadori2014fast,lozano2009spatial}. 
We used 19 variables for 1990, such as month, latitude, longitude, 
carbon dioxide and temperature, 
which were interpolated on a $2.5\times2.5$ degree grid with 125 locations. 
The second data set is the U.S. Historical Climatology Network (USHCN) data set 
\footnote{\url{https://www.ncdc.noaa.gov/oa/climate/research/ushcn/}},
which consists of monthly climate reports at 1218 locations in U.S. for 1990.
We used the following seven variables:
month, latitude, longitude, elevation, precipitation, 
minimum temperature, and maximum temperature.

The task was to estimate the distribution of a variable
given the values of the other variables as inputs;
there were 19 tasks in CC data, and seven tasks in USHCN data.
We evaluated the performance in terms of test log likelihoods.
We also used mean squared errors to evaluate point estimate performance.
We randomly selected some locations as test data.
The remaining data points were randomly split
into 90\% training data and 10\% validation data.
With CC data, we used 20\%, 50\% and 80\% of locations as test data,
and their training data sizes were 1081, 657 and 271, respectively.
With USHCN data, we used 50\%, 90\% and 95\% of locations as test data,
and their training data sizes were 6597, 1358 and 609, respectively.

\paragraph{Comparing Methods}

We compared the proposed method with GPs and NNs.
The GPs were sparse GPs inferred by stochastic variational inference.
The GPs correspond to the proposed method with a zero mean function.
With the proposed method and GPs, 
we used the following RBF kernels for the kernel function,
$k(\vec{x},\vec{x}')=\alpha \exp \left(-\frac{\gamma}{2}\parallel\vec{x}-\vec{x}'\parallel^{2}\right)$,
and 100 inducing points.
We set the step size at epoch $t$ as $\ell_{t}=(t+1)^{-0.9}$
for the stochastic variational inference.
With the NNs, we used three-layer feed-forward NNs with five hidden units,
and we optimized the NN parameters $\bm{\phi}$
and precision parameter $\beta$ 
by maximizing the following likelihood,
$\sum_{n=1}^{N}\log {\cal N}(y_{n}|g(\vec{x}_{n};\bm{\phi}),\beta^{-1})$,
by using Adam~\citep{kingma2014adam}.
The proposed method used NNs with the same structure 
for the mean function,
where the NN parameters were first optimized by maximizing the likelihood,
and then variational, kernel and NN parameters were estimated
by maximizing the variational lower bound (\ref{eq:lowerbound}) using the stochastic variational inference and Adam.
The locations of the inducing inputs were initialized by $k$-means results.
For all the methods, we set the minibatch size at 64,
and used early stoping based on the likelihood on a validation set. 

\paragraph{Results}

Tables~\ref{tab:likelihood_ccds} and \ref{tab:likelihood_ushcn}
show the test log likelihoods with different missing value rates
with CC and USHCN data, respectively.
The proposed method achieved the highest average likelihoods 
with both data sets.
The NN performed poorly when 
many values were missing (Table~\ref{tab:likelihood_ccds}, 80\% missing).
On the other hand, since a GP is a nonparametric Bayesian method,
where the effective model complexity is automatically adjusted
depending on the number of training samples,
the GPs performed better than the NNs
with the many missing value data.
When the number of missing values was small 
(Table~\ref{tab:likelihood_ccds}, 20\% missing),
the NN performed better than the GP.
The proposed method achieved the best performance
with different missing value rates by combining the advantages of NNs and GPs.
Tables~\ref{tab:mean_squared_error_ccds} 
and \ref{tab:mean_squared_error_ushcn} 
show the test mean squared errors with different missing value rates.
The proposed method achieved the lowest average errors with both data sets.
This result indicates that combining NNs and GPs also helps 
to improve the generalization performance.
Table~\ref{tab:time} shows the computational time in seconds.

\begin{figure}[t]
\centering
{\tabcolsep=-0.5em
\begin{tabular}{ccc}
\includegraphics[width=14.5em]{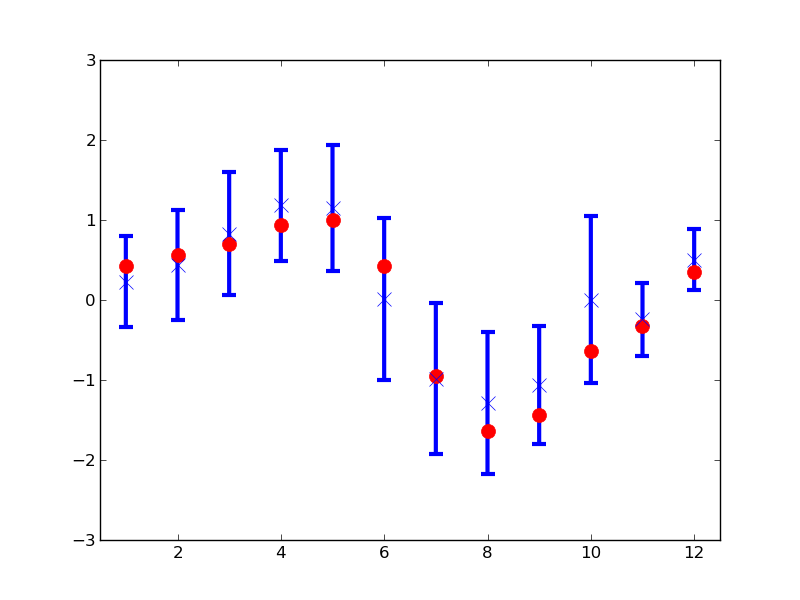}&
\includegraphics[width=14.5em]{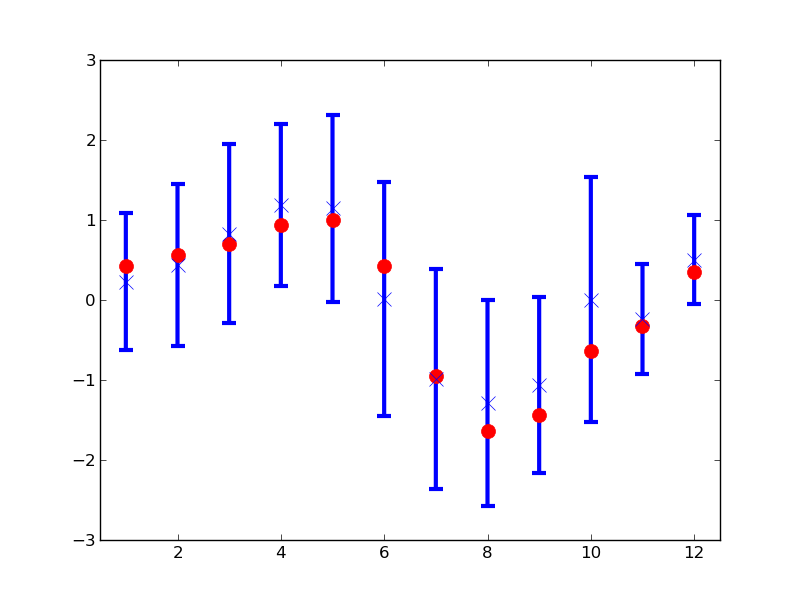}&
\includegraphics[width=14.5em]{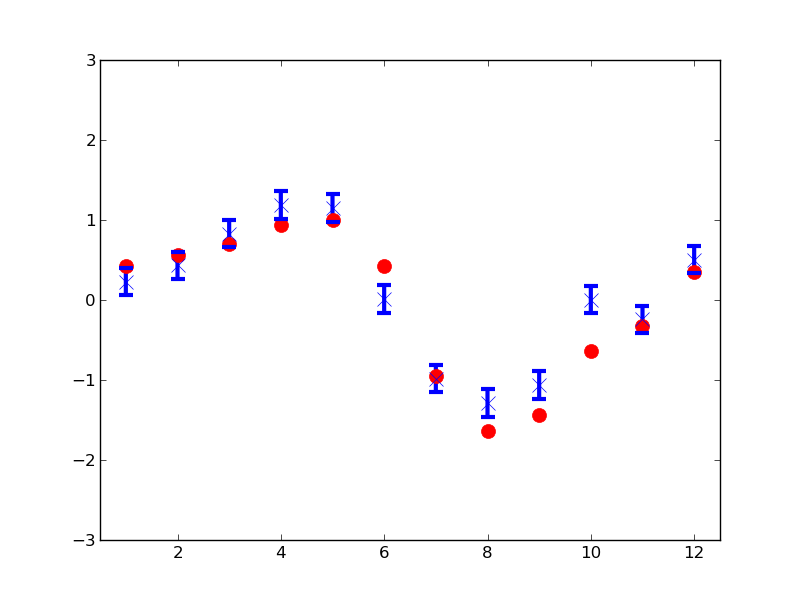}\\
(a) Proposed & (b) GP & (c) NN \\
\end{tabular}}
\caption{Prediction of held-out CO2 values using training data with 80\% missing values at a test location with CC data. The horizontal axis is month, the vertical axis is CO2, the blue bar is the 95\% confidence interval of the prediction, and the red point is the true value.}
\label{fig:predict_co2}
\end{figure}

Figure~\ref{fig:predict_co2} 
shows the prediction with its confidence interval obtained 
by the proposed method, GP and NN.
The NN gives fixed confidence intervals at all test points, 
and some true values are located outside the confidence intervals. 
On the other hand, the proposed method flexibly changes confidence intervals depending on the test points.
The confidence intervals with the GP differ across different test points as with the proposed method.
However, they are wider than those of the proposed method,
since the mean functions are fixed at zero.

\begin{table}[t]
\centering
\caption{Test log likelihoods provided by the proposed method, GP, and NN with CC data. The bottom row shows the values averaged over all variables. Values in a bold typeface are statistically better (at the 5\% level) than those in normal typeface as indicated by a paired t-test.}
\label{tab:likelihood_ccds}
{\tabcolsep=0.4em
\begin{tabular}{|l|rrr||rrr||rrr|}
\hline
Missing & \multicolumn{3}{|c||}{20\%} & \multicolumn{3}{c||}{50\%} & \multicolumn{3}{c|}{80\%} \\ 
\hline
Method  & Proposed  & GP  & NN  & Proposed  & GP  & NN  & Proposed  & GP  & NN  \\
\hline
\rotatebox{0}{ MON }   &  $\mathbf{1.75}$  &  $0.37$  &  $1.13$  &  $\mathbf{1.69}$  &  $0.20$  &  $1.09$  &  $\mathbf{-0.58}$  &  $\mathbf{-0.03}$  &  $\mathbf{-0.64}$  \\
 \rotatebox{0}{ LAT }   &  $\mathbf{1.85}$  &  $0.64$  &  $1.22$  &  $\mathbf{1.88}$  &  $0.63$  &  $1.18$  &  $\mathbf{0.88}$  &  $0.40$  &  $0.11$  \\
 \rotatebox{0}{ LON }   &  $\mathbf{-0.53}$  &  $\mathbf{-0.57}$  &  $-0.70$  &  $\mathbf{-0.66}$  &  $\mathbf{-0.65}$  &  $-0.77$  &  $\mathbf{-0.82}$  &  $\mathbf{-0.84}$  &  $-1.02$  \\
 \rotatebox{0}{ CO2 }   &  $\mathbf{1.47}$  &  $0.66$  &  $1.01$  &  $\mathbf{1.35}$  &  $0.45$  &  $0.88$  &  $\mathbf{0.70}$  &  $0.29$  &  $-0.14$  \\
 \rotatebox{0}{ CH4 }   &  $\mathbf{1.37}$  &  $0.87$  &  $0.76$  &  $\mathbf{1.09}$  &  $0.81$  &  $0.65$  &  $\mathbf{0.70}$  &  $0.51$  &  $0.23$  \\
 \rotatebox{0}{ CO }   &  $\mathbf{1.28}$  &  $0.30$  &  $0.82$  &  $\mathbf{1.32}$  &  $0.35$  &  $0.85$  &  $\mathbf{0.50}$  &  $0.11$  &  $-0.01$  \\
 \rotatebox{0}{ H2 }   &  $\mathbf{1.08}$  &  $0.42$  &  $0.64$  &  $\mathbf{0.92}$  &  $0.41$  &  $0.41$  &  $\mathbf{0.46}$  &  $\mathbf{0.26}$  &  $-0.36$  \\
 \rotatebox{0}{ WET }   &  $\mathbf{-0.59}$  &  $-0.68$  &  $-0.63$  &  $\mathbf{-0.60}$  &  $-0.69$  &  $-0.66$  &  $\mathbf{-0.86}$  &  $\mathbf{-0.76}$  &  $\mathbf{-0.96}$  \\
 \rotatebox{0}{ CLD }   &  $\mathbf{-0.30}$  &  $-0.39$  &  $-0.39$  &  $\mathbf{-0.33}$  &  $-0.38$  &  $-0.52$  &  $\mathbf{-0.56}$  &  $\mathbf{-0.51}$  &  $-0.71$  \\
 \rotatebox{0}{ VAP }   &  $\mathbf{0.53}$  &  $0.38$  &  $0.39$  &  $\mathbf{0.43}$  &  $0.32$  &  $0.13$  &  $\mathbf{0.08}$  &  $\mathbf{0.16}$  &  $-0.31$  \\
 \rotatebox{0}{ PRE }   &  $\mathbf{-0.74}$  &  $-0.90$  &  $\mathbf{-0.76}$  &  $\mathbf{-0.78}$  &  $-0.84$  &  $-0.82$  &  $\mathbf{-0.94}$  &  $\mathbf{-0.91}$  &  $\mathbf{-0.99}$  \\
 \rotatebox{0}{ FRS }   &  $\mathbf{0.55}$  &  $0.44$  &  $0.45$  &  $\mathbf{0.49}$  &  $\mathbf{0.43}$  &  $0.36$  &  $\mathbf{0.38}$  &  $\mathbf{0.33}$  &  $\mathbf{0.37}$  \\
 \rotatebox{0}{ DTR }   &  $\mathbf{2.70}$  &  $1.84$  &  $2.53$  &  $2.26$  &  $1.82$  &  $\mathbf{2.60}$  &  $\mathbf{1.78}$  &  $1.61$  &  $\mathbf{1.33}$  \\
 \rotatebox{0}{ TMN }   &  $\mathbf{3.98}$  &  $3.03$  &  $2.68$  &  $\mathbf{3.41}$  &  $3.02$  &  $2.19$  &  $\mathbf{2.77}$  &  $\mathbf{2.85}$  &  $2.00$  \\
 \rotatebox{0}{ TMP }   &  $\mathbf{3.76}$  &  $3.09$  &  $3.16$  &  $\mathbf{3.63}$  &  $3.05$  &  $2.99$  &  $2.74$  &  $\mathbf{2.92}$  &  $2.10$  \\
 \rotatebox{0}{ TMX }   &  $\mathbf{3.81}$  &  $3.06$  &  $3.13$  &  $\mathbf{3.76}$  &  $3.06$  &  $3.22$  &  $\mathbf{2.86}$  &  $\mathbf{2.88}$  &  $2.33$  \\
 \rotatebox{0}{ GLO }   &  $\mathbf{0.85}$  &  $0.78$  &  $0.78$  &  $\mathbf{0.76}$  &  $\mathbf{0.71}$  &  $0.67$  &  $\mathbf{0.51}$  &  $\mathbf{0.56}$  &  $\mathbf{0.36}$  \\
 \rotatebox{0}{ ETR }   &  $2.98$  &  $2.52$  &  $\mathbf{3.37}$  &  $2.95$  &  $2.48$  &  $\mathbf{3.26}$  &  $\mathbf{2.54}$  &  $2.24$  &  $\mathbf{2.45}$  \\
 \rotatebox{0}{ ETRN }   &  $\mathbf{3.12}$  &  $2.42$  &  $\mathbf{3.08}$  &  $2.78$  &  $2.34$  &  $\mathbf{3.08}$  &  $2.53$  &  $2.10$  &  $\mathbf{2.77}$  \\
\hline
 \rotatebox{0}{ Average }  &  $\mathbf{1.52}$ &  $0.96$ &  $1.19$ &  $\mathbf{1.39}$ &  $0.92$ &  $1.09$ &  $\mathbf{0.82}$ &  $\mathbf{0.75}$ &  $0.47$ \\
\hline
\end{tabular}}
\end{table}

\begin{table}[t]
\centering
\caption{Test log likelihoods provided by the proposed method, GP, and NN with USHCN data.}
\label{tab:likelihood_ushcn}
{\tabcolsep=0.4em
\begin{tabular}{|l|rrr||rrr||rrr|}
\hline
Missing & \multicolumn{3}{|c||}{50\%} & \multicolumn{3}{c||}{90\%} & \multicolumn{3}{c|}{95\%} \\ 
\hline
Method  & Proposed  & GP  & NN  & Proposed  & GP  & NN  & Proposed  & GP  & NN  \\
\hline
\rotatebox{0}{ MON }   &  $\mathbf{-1.33}$  &  $-1.39$  &  $\mathbf{-1.33}$  &  $\mathbf{-1.35}$  &  $-1.38$  &  $-1.36$  &  $\mathbf{-1.38}$  &  $\mathbf{-1.39}$  &  $-1.40$  \\
 \rotatebox{0}{ LAT }   &  $\mathbf{-0.73}$  &  $-1.04$  &  $\mathbf{-0.73}$  &  $\mathbf{-0.73}$  &  $-0.94$  &  $-0.76$  &  $\mathbf{-0.75}$  &  $-0.96$  &  $-0.78$  \\
 \rotatebox{0}{ LON }   &  $-1.12$  &  $-1.25$  &  $\mathbf{-1.10}$  &  $\mathbf{-1.14}$  &  $-1.19$  &  $\mathbf{-1.13}$  &  $\mathbf{-1.16}$  &  $-1.23$  &  $\mathbf{-1.16}$  \\
 \rotatebox{0}{ ELE }   &  $\mathbf{-1.16}$  &  $-1.27$  &  $\mathbf{-1.14}$  &  $\mathbf{-1.16}$  &  $-1.25$  &  $-1.19$  &  $\mathbf{-1.21}$  &  $\mathbf{-1.28}$  &  $-1.27$  \\
 \rotatebox{0}{ PRE }   &  $\mathbf{1.31}$  &  $0.93$  &  $1.23$  &  $\mathbf{1.27}$  &  $0.97$  &  $1.13$  &  $\mathbf{1.19}$  &  $0.89$  &  $0.91$  \\
 \rotatebox{0}{ TMIN }   &  $\mathbf{1.04}$  &  $0.90$  &  $0.83$  &  $\mathbf{0.98}$  &  $0.89$  &  $0.74$  &  $\mathbf{0.86}$  &  $0.81$  &  $0.68$  \\
 \rotatebox{0}{ TMAX }   &  $\mathbf{0.75}$  &  $0.26$  &  $0.71$  &  $\mathbf{0.71}$  &  $0.29$  &  $0.60$  &  $\mathbf{0.55}$  &  $0.25$  &  $0.45$  \\
\hline
 \rotatebox{0}{ Average }  &  $\mathbf{-0.18}$ &  $-0.41$ &  $-0.22$ &  $\mathbf{-0.20}$ &  $-0.37$ &  $-0.28$ &  $\mathbf{-0.27}$ &  $-0.41$ &  $-0.37$ \\
\hline
\end{tabular}}
\end{table}

\begin{table}[t]
\centering
\caption{Test mean squared error provided by the proposed method, GP, and NN with CC data.}
\label{tab:mean_squared_error_ccds}
{\tabcolsep=0.4em
\begin{tabular}{|l|rrr||rrr||rrr|}
\hline
Missing & \multicolumn{3}{|c||}{20\%} & \multicolumn{3}{c||}{50\%} & \multicolumn{3}{c|}{80\%} \\ 
\hline
Method  & Proposed  & GP  & NN  & Proposed  & GP  & NN  & Proposed  & GP  & NN  \\
\hline
\rotatebox{0}{ MON }   &  $\mathbf{0.002}$  &  $0.027$  &  $0.006$  &  $\mathbf{0.003}$  &  $0.029$  &  $0.006$  &  $\mathbf{0.012}$  &  $0.049$  &  $0.020$  \\
 \rotatebox{0}{ LAT }   &  $\mathbf{0.002}$  &  $0.013$  &  $0.008$  &  $\mathbf{0.002}$  &  $0.015$  &  $0.007$  &  $\mathbf{0.012}$  &  $0.026$  &  $0.036$  \\
 \rotatebox{0}{ LON }   &  $\mathbf{0.157}$  &  $\mathbf{0.163}$  &  $0.230$  &  $\mathbf{0.203}$  &  $\mathbf{0.204}$  &  $0.258$  &  $\mathbf{0.305}$  &  $\mathbf{0.320}$  &  $0.396$  \\
 \rotatebox{0}{ CO2 }   &  $\mathbf{0.003}$  &  $0.012$  &  $0.008$  &  $\mathbf{0.004}$  &  $0.018$  &  $0.010$  &  $\mathbf{0.014}$  &  $0.031$  &  $0.025$  \\
 \rotatebox{0}{ CH4 }   &  $\mathbf{0.004}$  &  $0.010$  &  $0.013$  &  $\mathbf{0.007}$  &  $0.012$  &  $0.015$  &  $\mathbf{0.016}$  &  $0.024$  &  $0.026$  \\
 \rotatebox{0}{ CO }   &  $\mathbf{0.005}$  &  $0.027$  &  $0.012$  &  $\mathbf{0.004}$  &  $0.027$  &  $0.011$  &  $\mathbf{0.028}$  &  $0.048$  &  $\mathbf{0.050}$  \\
 \rotatebox{0}{ H2 }   &  $\mathbf{0.007}$  &  $0.026$  &  $0.017$  &  $\mathbf{0.011}$  &  $0.031$  &  $0.023$  &  $\mathbf{0.030}$  &  $0.049$  &  $0.054$  \\
 \rotatebox{0}{ WET }   &  $\mathbf{0.189}$  &  $0.235$  &  $\mathbf{0.197}$  &  $\mathbf{0.196}$  &  $0.234$  &  $0.209$  &  $\mathbf{0.274}$  &  $\mathbf{0.256}$  &  $0.283$  \\
 \rotatebox{0}{ CLD }   &  $\mathbf{0.105}$  &  $0.125$  &  $0.115$  &  $\mathbf{0.120}$  &  $0.130$  &  $0.137$  &  $\mathbf{0.153}$  &  $\mathbf{0.151}$  &  $0.165$  \\
 \rotatebox{0}{ VAP }   &  $\mathbf{0.020}$  &  $0.027$  &  $0.024$  &  $\mathbf{0.026}$  &  $0.032$  &  $0.030$  &  $\mathbf{0.040}$  &  $\mathbf{0.038}$  &  $0.046$  \\
 \rotatebox{0}{ PRE }   &  $\mathbf{0.253}$  &  $0.327$  &  $\mathbf{0.257}$  &  $\mathbf{0.270}$  &  $0.300$  &  $0.280$  &  $\mathbf{0.345}$  &  $\mathbf{0.331}$  &  $\mathbf{0.377}$  \\
 \rotatebox{0}{ FRS }   &  $\mathbf{0.018}$  &  $0.025$  &  $0.022$  &  $\mathbf{0.021}$  &  $0.026$  &  $0.024$  &  $\mathbf{0.027}$  &  $0.032$  &  $\mathbf{0.028}$  \\
 \rotatebox{0}{ DTR }   &  $\mathbf{0.000}$  &  $0.002$  &  $\mathbf{0.000}$  &  $0.001$  &  $0.002$  &  $\mathbf{0.000}$  &  $\mathbf{0.002}$  &  $\mathbf{0.003}$  &  $\mathbf{0.002}$  \\
 \rotatebox{0}{ TMN }   &  $\mathbf{0.000}$  &  $0.000$  &  $0.000$  &  $\mathbf{0.000}$  &  $\mathbf{0.000}$  &  $0.001$  &  $\mathbf{0.000}$  &  $\mathbf{0.000}$  &  $0.001$  \\
 \rotatebox{0}{ TMP }   &  $\mathbf{0.000}$  &  $0.000$  &  $0.000$  &  $\mathbf{0.000}$  &  $0.000$  &  $0.000$  &  $0.000$  &  $\mathbf{0.000}$  &  $0.001$  \\
 \rotatebox{0}{ TMX }   &  $\mathbf{0.000}$  &  $0.000$  &  $0.000$  &  $\mathbf{0.000}$  &  $0.000$  &  $0.000$  &  $\mathbf{0.000}$  &  $\mathbf{0.000}$  &  $0.001$  \\
 \rotatebox{0}{ GLO }   &  $\mathbf{0.011}$  &  $0.013$  &  $0.012$  &  $\mathbf{0.013}$  &  $0.015$  &  $0.014$  &  $\mathbf{0.019}$  &  $\mathbf{0.019}$  &  $\mathbf{0.020}$  \\
 \rotatebox{0}{ ETR }   &  $0.000$  &  $0.000$  &  $\mathbf{0.000}$  &  $0.000$  &  $0.000$  &  $\mathbf{0.000}$  &  $\mathbf{0.000}$  &  $0.001$  &  $\mathbf{0.000}$  \\
 \rotatebox{0}{ ETRN }   &  $\mathbf{0.000}$  &  $0.001$  &  $\mathbf{0.000}$  &  $\mathbf{0.000}$  &  $0.001$  &  $\mathbf{0.000}$  &  $\mathbf{0.000}$  &  $0.001$  &  $\mathbf{0.000}$  \\
\hline
 \rotatebox{0}{ Average } &  $\mathbf{0.041}$ &  $0.054$ &  $0.048$ &  $\mathbf{0.046}$ &  $0.057$ &  $0.054$ &  $\mathbf{0.067}$ &  $0.073$ &  $0.081$ \\
\hline
\end{tabular}}
\end{table}

\begin{table}[t]
\centering
\caption{Test mean squared error provided by the proposed method, GP, and NN with USHCN data.}
\label{tab:mean_squared_error_ushcn}
{\tabcolsep=0.4em
\begin{tabular}{|l|rrr||rrr||rrr|}
\hline
Missing & \multicolumn{3}{|c||}{50\%} & \multicolumn{3}{c||}{90\%} & \multicolumn{3}{c|}{95\%} \\ 
\hline
Method  & Proposed  & GP  & NN  & Proposed  & GP  & NN  & Proposed  & GP  & NN  \\
\hline
 \rotatebox{0}{ MON }   &  $\mathbf{0.834}$  &  $0.936$  &  $\mathbf{0.838}$  &  $\mathbf{0.864}$  &  $0.924$  &  $0.880$  &  $\mathbf{0.925}$  &  $\mathbf{0.935}$  &  $\mathbf{0.938}$  \\
 \rotatebox{0}{ LAT }   &  $\mathbf{0.251}$  &  $0.457$  &  $\mathbf{0.253}$  &  $\mathbf{0.252}$  &  $0.383$  &  $0.265$  &  $\mathbf{0.262}$  &  $0.389$  &  $0.273$  \\
 \rotatebox{0}{ LON }   &  $0.547$  &  $0.717$  &  $\mathbf{0.524}$  &  $0.573$  &  $0.648$  &  $\mathbf{0.564}$  &  $\mathbf{0.603}$  &  $0.682$  &  $\mathbf{0.588}$  \\
 \rotatebox{0}{ ELE }   &  $\mathbf{0.581}$  &  $0.726$  &  $\mathbf{0.577}$  &  $\mathbf{0.586}$  &  $0.682$  &  $0.617$  &  $\mathbf{0.655}$  &  $\mathbf{0.808}$  &  $0.702$  \\
 \rotatebox{0}{ PRE }   &  $\mathbf{0.004}$  &  $0.012$  &  $0.005$  &  $\mathbf{0.005}$  &  $0.012$  &  $0.006$  &  $\mathbf{0.006}$  &  $0.013$  &  $0.007$  \\
 \rotatebox{0}{ TMIN }   &  $\mathbf{0.008}$  &  $0.012$  &  $0.011$  &  $\mathbf{0.009}$  &  $0.012$  &  $0.013$  &  $\mathbf{0.012}$  &  $0.014$  &  $0.014$  \\
 \rotatebox{0}{ TMAX }   &  $\mathbf{0.013}$  &  $0.042$  &  $0.014$  &  $\mathbf{0.015}$  &  $0.041$  &  $0.017$  &  $\mathbf{0.022}$  &  $0.044$  &  $\mathbf{0.022}$  \\
\hline
 \rotatebox{0}{ Average }   &  $\mathbf{0.320}$ &  $0.415$ &  $\mathbf{0.317}$ &  $\mathbf{0.329}$ &  $0.386$ &  $0.338$ &  $\mathbf{0.355}$ &  $0.412$ &  $0.364$ \\
\hline
\end{tabular}}
\end{table}

\begin{table}[t]
\centering
\caption{Computational time for inference in seconds.}
\label{tab:time}
\begin{tabular}{|l|rrr|rrr|}
\hline
Data & \multicolumn{3}{|c|}{CC} & \multicolumn{3}{|c|}{USHCN} \\ 
\hline
Missing & 20\% & 50\% & 80\% & 50\% & 90\% & 95\% \\
\hline
Proposed & $1409$ & $1020$ & $559$ & $5471$ & $1940$ & $1374$ \\
GP & $1163$ & $873$ & $463$ & $5074$ & $1854$ & $1388$ \\ 
NN & $334$ & $250$ & $109$ & $1746$ & $404$ & $226$ \\
\hline
\end{tabular}
\end{table}

\section{Conclusion}
\label{sec:conclusion}

In this paper, we proposed a simple method for combining neural networks
and Gaussian processes.
With the proposed method, neural networks are used as the mean function of Gaussian processes.
We present a scalable learning procedure based on stochastic gradient descent and stochastic variational inference.
With experiments using two real-world spatio-temporal data sets,
we demonstrated that the proposed method achieved better uncertainty estimation and generalization performance than neural networks and Gaussian processes.
There are several avenues that can be pursed as future work.
In our experiments, we used feed-forward neural networks.
We would like to use other types of neural networks, such as convolutional and recurrent neural networks.
Moreover, we plan to analyze the sensitivity with respect to 
the structure of the neural networks, the number of inducing points and the choice of kernels.
Finally, the mean function of neural networks 
could be inferred using Bayesian methods.

\bibliographystyle{abbrvnat}
\begin{small}
\bibliography{nips_2017}
\end{small}

\end{document}